\documentclass[final]{article}

\usepackage[nonatbib]{nips_2017}

\usepackage{nips_2017}

\usepackage[utf8]{inputenc} 
\usepackage[T1]{fontenc}    
\usepackage{hyperref}       
\usepackage{url}            
\usepackage{booktabs}       
\usepackage{amsfonts}       
\usepackage{nicefrac}       
\usepackage{microtype}      
\usepackage{algorithm}
\usepackage{algorithmic}
\usepackage{graphicx}
\usepackage{subfigure}
\usepackage{makecell}
\usepackage{multirow}
\usepackage{amsmath}
\usepackage{amssymb}
\usepackage{amsthm}
\usepackage{capt-of}

\newcolumntype{C}[1]{>{\centering\let\newline\\\arraybackslash\hspace{0pt}}m{#1}}

\def\B{{\bf B}}
\def\b{{\bf b}}

\def\X{{\bf X}}
\def\Y{{\bf Y}}

\def\Q{{\bf Q}}

\def\S{{\bf S}}
\def\x{{\bf x}}
\def\y{{\bf y}}
\def\z{{\bf z}}

\def\U{{\bf U}}
\def\u{{\bf u}}
\def\V{{\bf V}}
\def\v{{\bf v}}

\def\X{{\bf X}}
\def\z{{\bf z}}

\def\0{{\bf 0}}
\def\1{{\bf 1}}

\def\OM{{\mathcal O}}

\def\SM{{\mathcal S}}
\def\RB{{\mathbb R}}
\def\IB{{\mathbb I}}

\def\tr{\mathrm{tr}}

\title{Asymmetric Deep Supervised Hashing}

\author{
  Qing-Yuan Jiang, Wu-Jun Li\\
  National Key Laboratory for Novel Software Technology\\
  Collaborative Innovation Center of Novel Software Technology and Industrialization\\
  Department of Computer Science and Technology, Nanjing University, P. R. China \\
  \texttt{jiangqy@lamda.nju.edu.cn, liwujun@nju.edu.cn} \\
}

\begin{document}

\maketitle

\begin{abstract}
Hashing has been widely used for large-scale approximate nearest neighbor search because of its storage and search efficiency. Recent work has found that deep supervised hashing can significantly outperform non-deep supervised hashing in many applications. However, most existing deep supervised hashing methods adopt a symmetric strategy to learn one deep hash function for both query points and database~(retrieval) points. The training of these symmetric deep supervised hashing methods is typically time-consuming, which makes them hard to effectively utilize the supervised information for cases with large-scale database. In this paper, we propose a novel deep supervised hashing method, called \emph{\underline{a}symmetric \underline{d}eep \underline{s}upervised \underline{h}ashing~(ADSH)}, for large-scale nearest neighbor search. ADSH treats the query points and database points in an asymmetric way. More specifically, ADSH learns a deep hash function only for query points, while the hash codes for database points are directly learned. The training of ADSH is much more efficient than that of traditional symmetric deep supervised hashing methods. Experiments show that ADSH can achieve state-of-the-art performance in real applications.
\end{abstract}

\section{Introduction}
Approximate nearest neighbor~(ANN) search~\cite{DBLP:journals/cacm/AndoniI08,DBLP:conf/icml/Shrivastava014,DBLP:conf/nips/AndoniILRS15} has attracted much attention from machine learning community, with a lot of applications in information retrieval, computer vision and so on. As a widely used technique for ANN search, hashing~\cite{DBLP:conf/nips/WeissTF08,DBLP:conf/nips/KulisD09,DBLP:conf/icml/WangKC10,DBLP:conf/icml/NorouziF11,DBLP:conf/cvpr/GordoP11,DBLP:conf/nips/0002FS12,DBLP:conf/icml/LiuWKC11,DBLP:conf/nips/LiuMKC14,DBLP:conf/icml/Shrivastava014,DBLP:conf/nips/AndoniILRS15,DBLP:conf/cvpr/Carreira-Perpinan15,DBLP:conf/nips/Raziperchikolaei16} aims to encode the data points into compact binary hash codes. Due to its storage and search efficiency, hashing has attracted more and more attention for large-scale ANN search.

As the pioneering work, locality sensitive hashing~(LSH)~\cite{DBLP:conf/compgeom/DatarIIM04,DBLP:journals/cacm/AndoniI08} tries to utilize random projections as hash functions. LSH-like methods are always called data-independent methods, because the random projections are typically independent of training data. On the contrary, data-dependent methods~\cite{DBLP:conf/nips/LiuMKC14}, which are also called \emph{learning to hash~(L2H)} methods, aim to learn the hash functions from training data. Data-dependent methods usually achieve more promising performance than data-independent methods with shorter binary codes. Hence, data-dependent methods have become more popular than data-independent methods in recent years. Based on whether supervised information is used or not, data-dependent methods can be further divided into two categories~\cite{DBLP:conf/nips/LiuMKC14,DBLP:conf/ijcai/LiWK16}: unsupervised hashing and supervised hashing. Unsupervised hashing does not use supervised information for hash function learning. On the contrary, supervised hashing tries to learn the hash function by utilizing supervised information. In recent years, supervised hashing has attracted more and more attention because it can achieve better accuracy than unsupervised hashing~\cite{DBLP:conf/cvpr/LiuWJJC12,DBLP:conf/nips/NeyshaburSSMY13,DBLP:conf/ijcai/LiWK16}.

Most traditional supervised hashing methods are non-deep methods which cannot perform feature learning from scratch. Representative non-deep supervised hashing methods includes asymmetric hashing with two variants Lin:Lin and Lin:V~\cite{DBLP:conf/nips/NeyshaburSSMY13}, fast supervised hashing~(FastH)~\cite{DBLP:conf/cvpr/LinSSHS14}, supervised discrete hashing~(SDH)~\cite{DBLP:conf/cvpr/ShenSLS15} and column-sampling based discrete supervised hashing~(COSDISH)~\cite{DBLP:conf/aaai/KangLZ16}. Recently, deep supervised hashing, which adopts deep learning~\cite{DBLP:conf/nips/KrizhevskySH12} to perform feature learning for hashing, has been proposed by researchers. Representative deep supervised hashing methods include convolutional neural networks based hashing~(CNNH)~\cite{DBLP:conf/aaai/XiaPLLY14}, deep pairwise supervised hashing~(DPSH)~\cite{DBLP:conf/ijcai/LiWK16}, deep hashing network~(DHN)~\cite{DBLP:conf/aaai/ZhuL0C16} and deep supervised hashing~(DSH)~\cite{DBLP:conf/cvpr/Liu0SC16}. By integrating feature learning and hash-code learning into the same end-to-end architecture, deep supervised hashing can significantly outperform non-deep supervised hashing in many applications.

Most existing deep supervised hashing methods, including those mentioned above, adopt a symmetric strategy to learn one hash function for both query points and database points. The training of these symmetric deep supervised hashing methods is typically time-consuming. For example, the storage and computational cost for these hashing methods with pairwise supervised information is $\OM(n^2)$ where $n$ is the number of database points. The training cost for methods with triplet supervised information~\cite{DBLP:conf/cvpr/ZhaoHWT15} is even higher. To make the training practicable, most existing deep supervised hashing methods have to sample only a small subset from the whole database to construct a training set for hash function learning, and many points in database may be discarded during training. Hence, it is hard for these symmetric deep supervised hashing methods to effectively utilize the supervised information for cases with large-scale database, which makes the search performance unsatisfactory.

In this paper, we propose a novel deep supervised hashing method, called \emph{\underline{a}symmetric \underline{d}eep \underline{s}upervised \underline{h}ashing~(ADSH)}, for large-scale nearest neighbor search.  The main contributions of ADSH are outlined as follows: (1). ADSH treats the query points and database points in an asymmetric way. More specifically, ADSH learns a deep hash function only for query points, while the binary hash codes for database points are directly learned by adopting a bit by bit method. To the best of our knowledge, ADSH is the first deep supervised hashing method which treats query points and database points in an asymmetric way. (2). The training of ADSH is much more efficient than that of traditional symmetric deep supervised hashing methods. Hence, the whole set of database points can be used for training even if the database is large. (3). ADSH can directly learn the binary hash codes for database points, which will be empirically proved to be more accurate than the strategies adopted by traditional symmetric deep supervised hashing methods which use the learned hash function to generate hash codes for database points. (4). Experiments on two large-scale datasets show that ADSH can achieve state-of-the-art performance in real applications. 


\section{Notation and Problem Definition}
\label{sec:NoPro}
\subsection{Notation}
Boldface lowercase letters like $\b$ denote vectors, and boldface uppercase letters like $\B$ denote matrices. $\B_{*j}$ denotes the $j$th column of $\B$. $B_{ij}$ denotes the $(i,j)$th element of matrix $\B$. Furthermore, $\Vert\B\Vert_F$ and $\B^T$ are used to denote the Frobenius norm and the transpose of matrix $\B$, respectively. The symbol $\odot$ is used to denote the Hadamard product. We use $\IB(\cdot)$ to denote an indicator function, i.e., $\IB(true)=1$ and $\IB(false)=0$.

\subsection{Problem Definition}
For supervised hashing methods, supervised information can be point-wise labels~\cite{DBLP:conf/cvpr/ShenSLS15}, pairwise labels~\cite{DBLP:conf/aaai/KangLZ16,DBLP:conf/ijcai/LiWK16,DBLP:conf/cvpr/Liu0SC16} or triplet labels~\cite{DBLP:conf/iccv/WangLSJ13,DBLP:conf/cvpr/ZhaoHWT15}. In this paper, we only focus on pairwise-label based supervised hashing which is a common application scenario.

Assume that we have $m$ query data points which are denoted as $\X=\{\x_i\}_{i=1}^m$ and $n$ database points which are denoted as $\Y=\{\y_j\}_{j=1}^n$. Furthermore, pairwise supervised information, denoted as $\S\in\{-1,+1\}^{m\times n}$, is also available for supervised hashing. If $S_{ij}=1$, it means that point $\x_i$ and point $\y_j$ are similar. Otherwise, $\x_i$ and $\y_j$ are dissimilar. The goal of supervised hashing is to learn binary hash codes for both query points and database points from $\X$, $\Y$ and $\S$, and the hash codes try to preserve the similarity between query points and database points. More specifically, if we use $\U=\{\u_i\}_{i=1}^m\in\{-1,+1\}^{m\times c}$ and $\V=\{\v_j\}_{j=1}^n\in\{-1,+1\}^{n\times c}$ to denote the learned binary hash codes for query points and database points respectively, the Hamming distance $\text{dist}_H(\u_i,\v_j)$ should be as small as possible if $S_{ij}=1$ and vice versa. Here, $c$ denotes the binary code length. Moreover, we should also learn a hash function $h(\x_q)\in\{-1,+1\}^{c}$ so that we can generate binary code for any unseen query point $\x_q$.

Please note that in many cases, we are only given a set of database points $\Y=\{\y_j\}_{j=1}^n$ and the pairwise supervised information between them. We can also learn the hash codes and hash function by sampling a subset or the whole set of $\Y$ as the query set for training. In these cases, $\X \subseteq\Y$.


\begin{figure*}[t]
\centering
\includegraphics[width=0.85\linewidth]{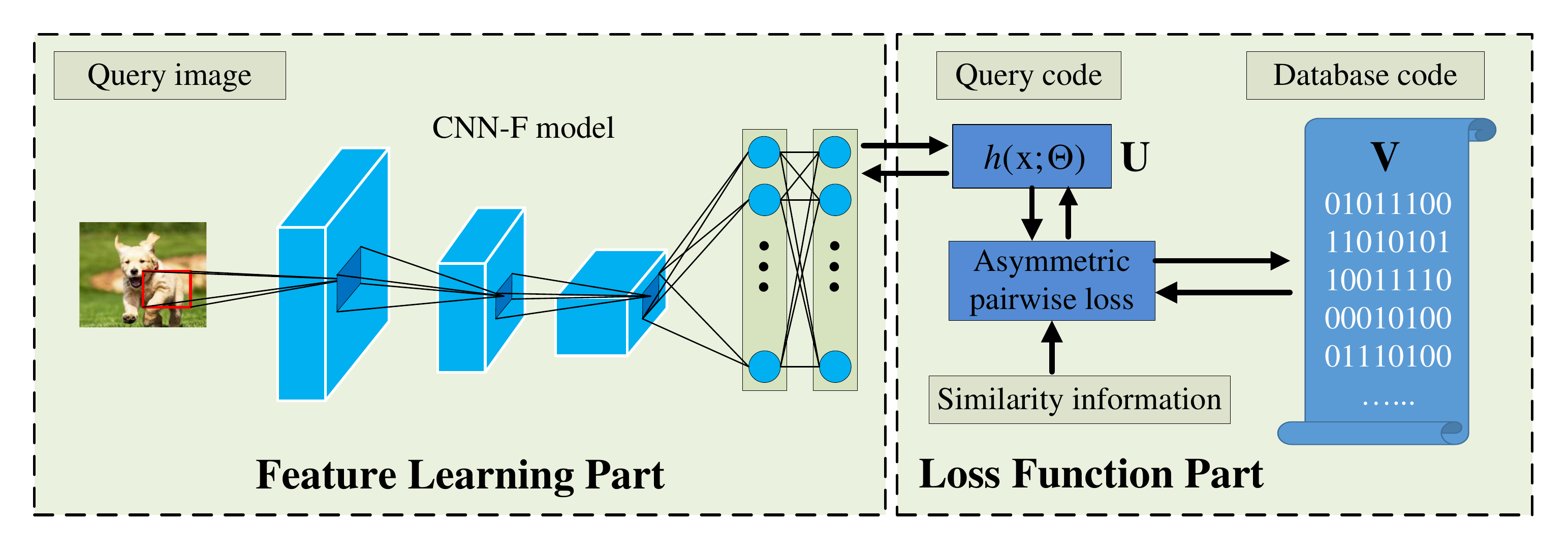}
\caption{Model architecture of ADSH.}
\label{fig:framework}
\end{figure*}

\section{Asymmetric Deep Supervised Hashing}
\label{sec:ADSH}

In this section, we introduce our asymmetric deep supervised hashing~(ADSH) in detail, including model formulation and learning algorithm.
\subsection{Model Formulation}

Figure~\ref{fig:framework} shows the model architecture of ADSH, which contains two important components: \emph{feature learning part} and \emph{loss function part}. The feature learning part tries to learn a deep neural network which can extract appropriate feature representation for binary hash code learning. The loss function part aims to learn binary hash codes which preserve the supervised information~(similarity) between query points and database points. ADSH integrates these two components into the same end-to-end framework. During training procedure, each part can give feedback to the other part.

Please note that the feature learning is only performed for query points but not for database points. Furthermore, based on the deep neural network for feature learning, ADSH adopts a deep hash function to generate hash codes for query points, but the binary hash codes for database points are directly learned. Hence, ADSH treats the query points and database points in an asymmetric way.  This asymmetric property of ADSH is different from traditional deep supervised hashing methods. Traditional deep supervised hashing methods adopt the same deep neural network to perform feature learning for both query points and database points, and then use the same deep hash function to generate binary codes for both query points and database points.


\subsubsection{Feature Learning Part}
In this paper, we adopt a convolutional neural network~(CNN) model from~\cite{DBLP:conf/bmvc/ChatfieldSVZ14}, i.e., CNN-F model, to perform feature learning. This CNN-F model has also been adopted in DPSH~\cite{DBLP:conf/ijcai/LiWK16} for feature learning. The CNN-F model contains five convolutional layers and three fully-connected layers, the details of which can be found at~\cite{DBLP:conf/bmvc/ChatfieldSVZ14,DBLP:conf/ijcai/LiWK16}. In ADSH, the last layer of CNN-F model is replaced by a fully-connected layer which can project the output of the first seven layers into $\RB^c$ space. Please note that the framework of ADSH is general enough to adopt other deep neural networks to replace the CNN-F model for feature learning. Here, we just adopt CNN-F for illustration.
\subsubsection{Loss Function Part}
To learn the hash codes which can preserve the similarity between query points and database points, one common way is to minimize the $L_2$ loss between the supervised information~(similarity) and inner product of query-database binary code pairs. This can be formulated as follows:
\begin{align}\label{pro:obj}
\min_{\U,\V}\; & J(\U, \V)=\sum_{i=1}^m\sum_{j=1}^n\big(\u_i^T\v_j-cS_{ij}\big)^2\\
\text{s.t.}\quad&\U\in\{-1,+1\}^{m\times c},\V\in\{-1,+1\}^{n\times c},\u_i=h(\x_i)\;, \forall i\in\{1,2,\dots,m\}. \nonumber
\end{align}

However, it is difficult to learn $h(\x_i)$ due to the discrete output. We can set $h(\x_i)=\text{sign}(F(\x_i;\Theta))$, where $F(\x_i;\Theta) \in \RB^c$. Then, the problem in~(\ref{pro:obj}) is transformed to the following problem:
\begin{align}\label{pro:reobj}
\min_{\Theta,\V}\; J(\Theta,\V)=\sum_{i=1}^m\sum_{j=1}^n\big[\text{sign}(F(\x_i;\Theta))^T\v_j-cS_{ij}\big]^2,\; \hspace{0.5cm}
\text{s.t.}\quad\V\in\{-1,+1\}^{n\times c}.
\end{align}
In~(\ref{pro:reobj}), we set $F(\x_i;\Theta)$ to be the output of the CNN-F model in the feature learning part, and $\Theta$ is the parameter of the CNN-F model. Through this way, we seamlessly integrate the feature learning part and the loss function part into the same framework.

There still exists a problem for the formulation in~(\ref{pro:reobj}), which is that we cannot back-propagate the gradient to $\Theta$ due to the $\text{sign}(F(\x_i;\Theta))$ function. Hence, in ADSH we adopt the following objective function:
\begin{align}\label{pro:reobjADSH}
\min_{\Theta,\V}\; J(\Theta,&\V)=\sum_{i=1}^m\sum_{j=1}^n\big[\text{tanh}(F(\x_i;\Theta))^T\v_j-cS_{ij}\big]^2,\; \hspace{0.5cm}\text{s.t.}\quad \V\in\{-1,+1\}^{n\times c}
\end{align}
where we use $\text{tanh}(\cdot)$ to approximate the $\text{sign}(\cdot)$ function.


In practice, we might be given only a set of database points $\Y=\{\y_j\}_{j=1}^n$ without query points. In this case, we can randomly sample $m$ data points from database to construct the query set. More specifically, we set $\X=\Y^\Omega$ where $\Y^\Omega$ denotes the database points indexed by $\Omega$. Here, we use $\Gamma=\{1,2,\dots,n\}$ to denote the indices of all the database points and $\Omega=\{i_1,i_2,\dots,i_m\}\subseteq\Gamma$ to denote the indices of the sampled query points. Accordingly, we set $\S=\SM^\Omega$, where $\SM\in\{-1,+1\}^{n\times n}$ denotes the supervised information~(similarity) between pairs of all database points and $\SM^\Omega\in\{-1,+1\}^{m\times n}$ denotes the sub-matrix formed by the rows of $\SM$ indexed by $\Omega$. Then, we can rewrite $J(\Theta,\V)$ as: $J(\Theta,\V)=\sum_{i\in\Omega}\sum_{j\in\Gamma}\big[\text{tanh}(F(\y_i;\Theta))^T\v_j-cS_{ij}\big]^2$.

Because $\Omega \subseteq \Gamma$, there are two representations for $\y_i$, $\forall i\in\Omega$. One is the binary hash code $\v_i$ in database, and the other is the query representation $\text{tanh}(F(\y_i;\Theta))$. We add an extra constraint to keep $\v_i$ and $\text{tanh}(F(\y_i;\Theta))$ as close as possible, $\forall i\in\Omega$. This is intuitively reasonable, because $\text{tanh}(F(\y_i;\Theta))$ is the approximation of the binary code of $\y_i$. Then we get the final formulation of ADSH with only database points $\Y$ for training:
\begin{align}
\min_{\Theta,\V}\; J(\Theta,\V)&=\sum_{i\in\Omega}\sum_{j\in\Gamma}\big[\text{tanh}(F(\y_i;\Theta))^T\v_j-cS_{ij}\big]^2+\gamma\sum_{i\in\Omega}[\v_i-\text{tanh}(F(\y_i;\Theta))]^2\nonumber\\
\text{s.t.}\quad &\;\V\in\{-1,+1\}^{n\times c}
\label{pro:objADSH}
\end{align}
where $\gamma$ is a hyper-parameter.

In real applications, if we are given both $\Y$ and $\X$, we use the problem in~(\ref{pro:reobjADSH}) for training ADSH. If we are only given $\Y$, we use the problem in~(\ref{pro:objADSH}) for training ADSH. After training ADSH, we can get the binary hash codes for database points, and a deep hash function for query points. We can use the trained deep hash function to generate the binary hash codes for any query points including newly coming query points which are not seen during training. One simple way to generate binary codes for query points is to set $\u_q = h(\x_q) = \text{sign}(F(\x_q;\Theta))$.

From (\ref{pro:reobjADSH}) and (\ref{pro:objADSH}), we can find that ADSH treats query points and database points in an asymmetric way. More specifically, the feature learning is only performed for query points but not for database points. Furthermore, ADSH adopts a deep hash function to generate hash codes for query points, but the binary hash codes for database points are directly learned. This is different from traditional deep supervised hashing methods which adopt the same deep hash function to generate binary hash codes for both query points and database points. Because $m\ll n$ in general, ADSH can learn the deep neural networks efficiently, and is much faster than traditional symmetric deep supervised hashing methods. This will be verified in our experiments.

\subsection{Learning Algorithm}
Here, we only present the learning algorithm for problem~(\ref{pro:objADSH}), which can be easily adapted for problem~(\ref{pro:reobjADSH}). We design an alternating optimization strategy to learn the parameters $\Theta$ and $\V$ in problem~(\ref{pro:objADSH}). More specifically, in each iteration we learn one parameter with the other fixed, and this process will be repeated for many iterations.

\subsubsection{Learn $\Theta$ with $\V$ fixed}
When $\V$ is fixed, we use back-propagation~(BP) algorithm to update the neural network parameter $\Theta$. Specifically, we sample a mini-batch of the query points, then update the parameter $\Theta$ based on the sampled data. For the sake of simplicity, we define $\z_i=F(\y_i;\Theta)$ and $\widetilde\u_i=\text{tanh}(F(\y_i;\Theta))$. Then we can compute the gradient of $\z_i$ as follows:
\begin{align}
\frac{\partial J}{\partial\z_i} = &2\Big\{\sum_{j\in\Gamma}\big[(\widetilde\u_i^T\v_j-c S_{ij})\v_j\big]+2\gamma(\widetilde\u_i-\v_i)\Big\}\odot(1-\widetilde\u_i^2)
\label{pro:gradv}
\end{align}
Then we can use chain rule to compute $\frac{\partial J}{\partial \Theta}$ based on $\frac{\partial J}{\partial \z_i}$, and the BP algorithm is used to update $\Theta$.

\subsubsection{Learn $\V$ with $\Theta$ fixed}
When $\Theta$ is fixed, we rewrite the problem~(\ref{pro:objADSH}) in matrix form:
\begin{align}\label{pro:objU}
\min_{\V}\;J(\V)=\Vert\widetilde\U\V^T-c\S\Vert^2_F+\gamma\Vert\V^\Omega-\widetilde\U\Vert^2_F \hspace{0.5cm} \text{s.t.}\quad\V\in\{-1,+1\}^{n\times c},
\end{align}
where $\widetilde\U=[\widetilde\u_{i_1},\widetilde\u_{i_2},\dots,\widetilde\u_{i_m}]^T\in[-1,+1]^{m\times c}$, $\V^\Omega$ denotes the binary codes for the database points indexed by $\Omega$, i.e., $\V^\Omega=[\v_{i_1},\v_{i_2},\dots,\v_{i_m}]^T$.

We define $\bar\U=\{\bar\u_j\}_{j=1}^n$, where $\bar\u_j$ is defined as: $\bar\u_{j}=\IB(j\in\Omega)\cdot\widetilde\u_{j}+\IB(j\notin\Omega)\cdot{\bf 0}$. Then we can rewrite the problem~(\ref{pro:objU}) as follows:
\begin{align}
\min_{\V}\;J(\V)=&\Vert\V\widetilde\U^T\Vert^2_F-2\tr\big(\V[c\widetilde\U^T\S+\gamma\bar\U^T]\big)+\text{const}=\Vert\V\widetilde\U^T\Vert^2_F+\tr(\V\Q^T)+\text{const}\nonumber\\
\text{s.t.}\quad&\V\in\{-1,+1\}^{n\times c}
\label{pro:objUk}
\end{align}
where $\Q=-2c\S^T\widetilde\U-2\gamma\bar\U$,  const is a constant independent of $\V$.

Then, we update $\V$ bit by bit. That is to say, each time we update one column of $\V$ with other columns fixed. Let $\V_{*k}$ denote the $k$th column of $\V$ and $\widehat\V_{k}$ denote the matrix of $\V$ excluding $\V_{*k}$. Let $\Q_{*k}$ denote the $k$th column of $\Q$ and $\widehat\Q_{k}$ denote the matrix of $\Q$ excluding $\Q_{*k}$.  Let $\widetilde\U_{*k}$ denote the $k$th column of $\widetilde\U$ and $\widehat\U_{k}$ denote the matrix of $\widetilde\U$ excluding $\widetilde\U_{*k}$. To optimize $\V_{*k}$, we can get the objective function: $J(\V_{*k})=\Vert\V\widetilde\U^T\Vert^2_F+\tr(\V\Q^T)+\text{const}=\tr\big(\V_{*k}[2\widetilde\U_{*k}^T\widehat\U_k\widehat\V_k^T+\Q^T_{*k}]\big)+\text{const}$. Then, we need to solve the following problem:
\begin{align}
\min_{\V_{*k}}\; J(\V_{*k})=\tr(\V_{*k}[2\widetilde\U_{*k}^T\widehat\U_k\widehat\V_k^T+\Q^T_{*k}]\big)+\text{const}\hspace{0.5cm}\text{s.t.}\quad\V_{*k}\in\{-1,+1\}^{n}.
\label{pro:Uk}
\end{align}
Then, we can get the optimal solution of problem~(\ref{pro:Uk}) as follows: 
\begin{align}
\V_{*k}=-\text{sign}(2\widehat\V_k\widehat\U_k^T\widetilde\U_{*k}+\Q_{*k}), \label{eq:updateV}
\end{align}
which can be used to update $\V_{*k}$.

We summarize the whole learning algorithm for ADSH in Algorithm~\ref{alg:ADSH}. Here, we repeat the learning for several times, and each time we can sample a query set indexed by $\Omega$.
\begin{algorithm}[!htb]
\caption{The learning algorithm for ADSH}
\label{alg:ADSH}
\begin{algorithmic}
\REQUIRE $\Y=\{\y_i\}_{i=1}^n$: $n$ data points. $\SM\in\{-1,1\}^{n\times n}$: similarity matrix. $c$: code length.
\ENSURE  $\V$: binary hash codes for database points. $\Theta$: neural network parameter.
\STATE {\bf Initialization}: initialize $\Theta$, $\V$, mini-batch size $M$ and iteration number $T_{out},T_{in}$.
\FOR {$w=1\to T_{out}$}
    \STATE Randomly generate index set $\Omega$ from $\Gamma$. Set $\S=\SM^{\Omega}$, $\X=\Y^\Omega$ based on $\Omega$.
    \FOR {$t=1\to T_{in}$}
        \FOR {$s=1,2,\dots,m/M$}
            \STATE Randomly sample $M$ data points from $\X=\Y^\Omega$ to construct a mini-batch.
            \STATE Calculate $\z_i$ and $\widetilde\u_i$ for each data point $\y_i$ in the mini-batch by forward propagation.
            \STATE Calculate the gradient according to~(\ref{pro:gradv}).
            \STATE Update the parameter $\Theta$ by using back propagation.
        \ENDFOR
        \FOR {$k=1\to c$}
            \STATE Update $\V_{*k}$ according to update rule in~(\ref{eq:updateV}).
        \ENDFOR
    \ENDFOR
\ENDFOR
\end{algorithmic}
\end{algorithm}

\subsection{Out-of-Sample Extension}
After training ADSH, the learned deep neural network can be applied for generating binary codes for query points including unseen query points during training. More specifically, we can use the equation: $\u_q=h(\x_q;\Theta)=\text{sign}(F(\x_q;\Theta))$,
 to generate binary code for $\x_q$.

\subsection{Complexity Analysis}
The total computational complexity for training ADSH is $\OM(T_{out}T_{in}[(n+2)mc+(m+1)nc^2+(c(n+m)-m)c])$. In practice, $T_{out}$, $T_{in}$, $m$ and $c$ will be much less than $n$. Hence, the computational cost of ADSH is $\OM(n)$. For traditional symmetric deep supervised hashing methods, if all database points are used for training, the computational cost is at least $\OM(n^2)$. Furthermore, the training for deep neural network is typically time-consuming. For traditional symmetric deep supervised hashing methods, they need to scan $n$ points in an epoch of the neural network training. On the contrary, only $m$ points are scanned in an epoch of the neural network training for ADSH. Typically, $m\ll n$. Hence, ADSH is much faster than traditional symmetric deep supervised hashing methods.

To make the training practicable, most existing symmetric deep supervised hashing methods have to sample only a small subset from the whole database to construct a training set for deep hash function learning, and many points in database may be discarded during training. On the contrary, ASDH is much more efficient to utilize more database points for training.

\section{Experiment}
\label{sec:Exp}

We carry out experiments to evaluate our ADSH and baselines which are implemented with the deep learning toolbox MatConvNet~\cite{DBLP:conf/mm/VedaldiL15} on a NVIDIA K40 GPU server.

\subsection{Datasets}

We evaluate ADSH on two widely used datasets: \mbox{CIFAR-10}~\cite{krizhevsky2009learning} and \mbox{NUS-WIDE}~\cite{DBLP:conf/civr/ChuaTHLLZ09}.

The CIFAR-10 dataset is a single-label dataset which contains 60,000 32 $\times$ 32 color images. Each image belongs to one of the ten classes. Two images will be treated as a ground-truth neighbor~(similar pair) if they share one common label.

The NUS-WIDE dataset is a multi-label dataset which consists of 269,648 web images associated with tags. Following the setting of DPSH~\cite{DBLP:conf/ijcai/LiWK16}, we only select 195,834 images that belong to the 21 most frequent concepts. For NUS-WIDE, two images will be defined as a ground-truth neighbor~(similar pair) if they share at least one common label.

\subsection{Baselines and Evaluation Protocol}
\label{subsec:baselines}
To evaluate our ADSH, ten start-of-the-art methods, including ITQ~\cite{DBLP:conf/cvpr/GongL11}, Lin:Lin~\cite{DBLP:conf/nips/NeyshaburSSMY13}, Lin:V~\cite{DBLP:conf/nips/NeyshaburSSMY13}, LFH~\cite{DBLP:conf/sigir/ZhangZLG14}, FastH~\cite{DBLP:conf/cvpr/LinSSHS14}, SDH~\cite{DBLP:conf/cvpr/ShenSLS15}, COSDISH~\cite{DBLP:conf/aaai/KangLZ16}, DSH~\cite{DBLP:conf/cvpr/Liu0SC16}, DHN~\cite{DBLP:conf/aaai/ZhuL0C16} and DPSH~\cite{DBLP:conf/ijcai/LiWK16}, are selected as baselines for comparison.

For non-deep hashing methods, we utilize 4,096-dim deep features which are extracted by the pre-trained CNN-F model on ImageNet dataset for fair comparison. For FastH, LFH and COSDISH, we use boosted decision tree for out-of-sample extension by following the setting of FastH. For DSH and DHN, although the authors provide source code, for fair comparison we carefully re-implement their methods on MatConvNet to remove effect on training time caused by different platforms. For deep hashing methods, we resize all images to 224 $\times$ 224 and use the raw pixels as the inputs for all datasets. In addition, some deep hashing methods adopt other neural networks for feature learning. We find that the deep baselines with CNN-F network can outperform the counterparts with the original networks~(refer to Appendix C in the supplementary material). For fair comparison, we adopt the same deep neural networks for all deep hashing methods, i.e., the CNN-F network. We initialize CNN-F with the pre-trained model on ImageNet. Following the suggestions of the authors, we set the mini-batch size to be 128 and tune the learning rate among $[10^{-6},10^{-2}]$. For ADSH method, we set $\gamma=200$, $T_{out}=50$ by using a validation strategy. Furthermore, we set $T_{in}=3,5$, $\vert\Omega\vert=1000, 2000$ for CIFAR-10 and NUS-WIDE, respectively~\footnote{We report the effect of hyper-parameters $\gamma$ and $\vert\Omega\vert$ in Appendix D of the supplementary materials.}. To avoid effect caused by class-imbalance problem between positive and negative similarity information, inspired by~\cite{DBLP:conf/cikm/LengCWZL14}, we empirically set the weight of the element -1 in $\S$ as the ratio between the number of element 1 and the number of element -1 in $\S$.

For CIFAR-10 dataset, we randomly select 1,000 images for validation set and 1,000 images for test set, with the remaining images as database points. For NUS-WIDE dataset, we randomly choose 2,100 images as validation set and 2,100 images as test set, with the rest of the images as database points. Because the deep hashing baselines are very time-consuming for training, similar to existing works like~\cite{DBLP:conf/ijcai/LiWK16} we randomly sample 5,000~(500 per class) and 10,500 images from database for training all baselines except Lin:V for CIFAR-10 and NUS-WIDE, respectively. The necessity of random sampling for training set will also be empirically verified in Section~\ref{sec:timeExp}.  

We report Mean Average Precision~(MAP), Top-k precision curve to evaluate ADSH and baselines. For NUS-WIDE dataset, the MAP results are calculated based on the top-5000 returned samples. We also compare the training time between different deep hashing methods. Furthermore, we also report the precision-recall curve and case study, which are moved to the supplementary material due to the space limitation. All experiments are run five times, and the average values are reported.

\subsection{Accuracy}
The MAP results are presented in Table~\ref{table:map}. We can find that in most cases the supervised methods can outperform the unsupervised methods, and the deep methods can outperform the non-deep methods. Furthermore, we can find that ADSH can outperform all the other baselines, including deep hashing baselines, non-deep supervised hashing baselines and unsupervised hashing baselines.

\begin{table*}[tb]
\centering\small
\caption{MAP comparison to baselines. The best accuracy is shown in boldface.}
\label{table:map}
\begin{tabular}{|c||c|c|c|c||c|c|c|c|c|}
 \hline
 \multirow{2}{*}{Method} &
 \multicolumn{4}{c||}{{CIFAR-10}} & \multicolumn{4}{c|}{{NUS-WIDE}}\\
 \cline{2-9} & 12 bits & 24 bits & 32 bits & 48 bits & 12 bits & 24 bits & 32 bits & 48 bits\\
 \hline \hline
 ITQ & {0.2611} &{0.2734} &{0.2865} &{0.2957} & {0.6785} &{0.7020} &{0.7104} &{0.7204}
\\
 \Xcline{1-9}{1pt}
 Lin:Lin & {0.6173} &{0.6329} &{0.6166} &{0.5918} & {0.6464} &{0.6446} &{0.6278} &{0.6295}\\
 \hline
 FastH & {0.6111} &{0.6713} &{0.6874} &{0.7065} & {0.7506} &{0.7785} &{0.7853} &{0.7993}\\
 \hline
 LFH & {0.4342} &{0.5988} &{0.6551} &{0.6745} & {0.7362} &{0.7760} &{0.7896} &{0.8020}\\
 \hline
 SDH & {0.4866} &{0.6328} &{0.6457} &{0.6539} & {0.7550} &{0.7723} &{0.7727} &{0.7830}\\
 \hline
 COSDISH &{0.6048} &{0.6684} &{0.6871} &{0.7068}& {0.7418} &{0.7738} &{0.7895} &{0.7986}\\
 \Xcline{1-9}{1pt}
 DSH & {0.6370} &{0.7493} &{0.7803} &{0.8086} & {0.7650} &{0.7778} &{0.7822} &{0.7844}\\
 \hline
 DHN & {0.6794} &{0.7201} &{0.7309} &{0.7408} & {0.7598} &{0.7894} &{0.7960} &{0.8021}\\
 \hline
 DPSH & {0.6863} &{0.7276} &{0.7406} &{0.7520} & {0.7895} &{0.8083} &{0.8138} &{0.8216}\\
 \Xcline{1-9}{1pt}
 ADSH & {\bf 0.8466} &{\bf 0.9062} &{\bf 0.9175} &{\bf 0.9263}& {\bf 0.8570} &{\bf 0.8939} &{\bf 0.9008} &{\bf 0.9074}\\
 \hline
 \end{tabular}
\end{table*}


Some baselines, including Lin:Lin, LFH, SDH, COSDISH, can also be adapted to learn binary hash codes for database directly due to their training efficiency. We also carry out experiments to evaluate the adapted counterparts of these methods which can learn binary codes for database directly, and denote the counterparts of these methods as Lin:V, LFH-D, SDH-D, COSDISH-D, respectively. It also means that Lin:V, LFH-D, SDH-D, COSDISH-D adopt all database points for training. We report the corresponding MAP results in Table~\ref{table:mapBd}. We can find that Lin:V, LFH-D, SDH-D, COSDISH-D can outperform Lin:Lin, LFH, SDH, COSDISH, respectively. This means that directly learning the binary hash codes for database points is more accurate than the strategies which use the learned hash function to generate hash codes for database points. We can also find that ADSH can outperform all the other baselines.

\begin{table*}[tb]
\centering\small
\caption{MAP comparison to baselines which can directly learn the binary hash codes for database points. The best accuracy is shown in boldface.}
\label{table:mapBd}
\begin{tabular}{|c||c|c|c|c||c|c|c|c|c|}
 \hline
 \multirow{2}{*}{Method} &
 \multicolumn{4}{c||}{{CIFAR-10}} & \multicolumn{4}{c|}{{NUS-WIDE}}\\
 \cline{2-9} & 12 bits & 24 bits & 32 bits & 48 bits & 12 bits & 24 bits & 32 bits & 48 bits\\
 \hline \hline
 Lin:V & {0.8197} &{0.8011} &{0.8055} &{0.8021} & {0.7762} &{0.7756} &{0.7806} &{0.7820}\\
 \hline
 LFH-D & {0.4743} &{0.6384} &{0.7598} &{0.8142} & {0.7954} &{0.8393} &{0.8584} &{0.8721}\\
 \hline
 SDH-D & {0.7115} &{0.8297} &{0.8314} &{0.8336}& {0.8478} &{0.8683} &{0.8647} &{0.8670}\\
 \hline
 COSDISH-D &{0.8431} &{0.8666} &{0.8738} &{0.8724} &{0.8031} &{0.8448} &{0.8581} &{0.8670} \\
 \Xcline{1-9}{1pt}
 ADSH & {\bf 0.8466} &{\bf 0.9062} &{\bf 0.9175} &{\bf 0.9263}& {\bf 0.8570} &{\bf 0.8939} &{\bf 0.9008} &{\bf 0.9074}\\
 \hline
 \end{tabular}
\end{table*}


We also report top-2000 precision in Figure~\ref{fig:top2kpre} on two datasets. Once again, we can find that ADSH can significantly outperform other baselines in all cases especially for large code length.

\subsection{Time Complexity}\label{sec:timeExp}
\begin{figure}[t]
\minipage{0.49\textwidth}
\begin{tabular}{c@{}@{}c@{}@{}c@{}@{}c}
\begin{minipage}{0.49\linewidth}\centering
    \includegraphics[width=1\textwidth]{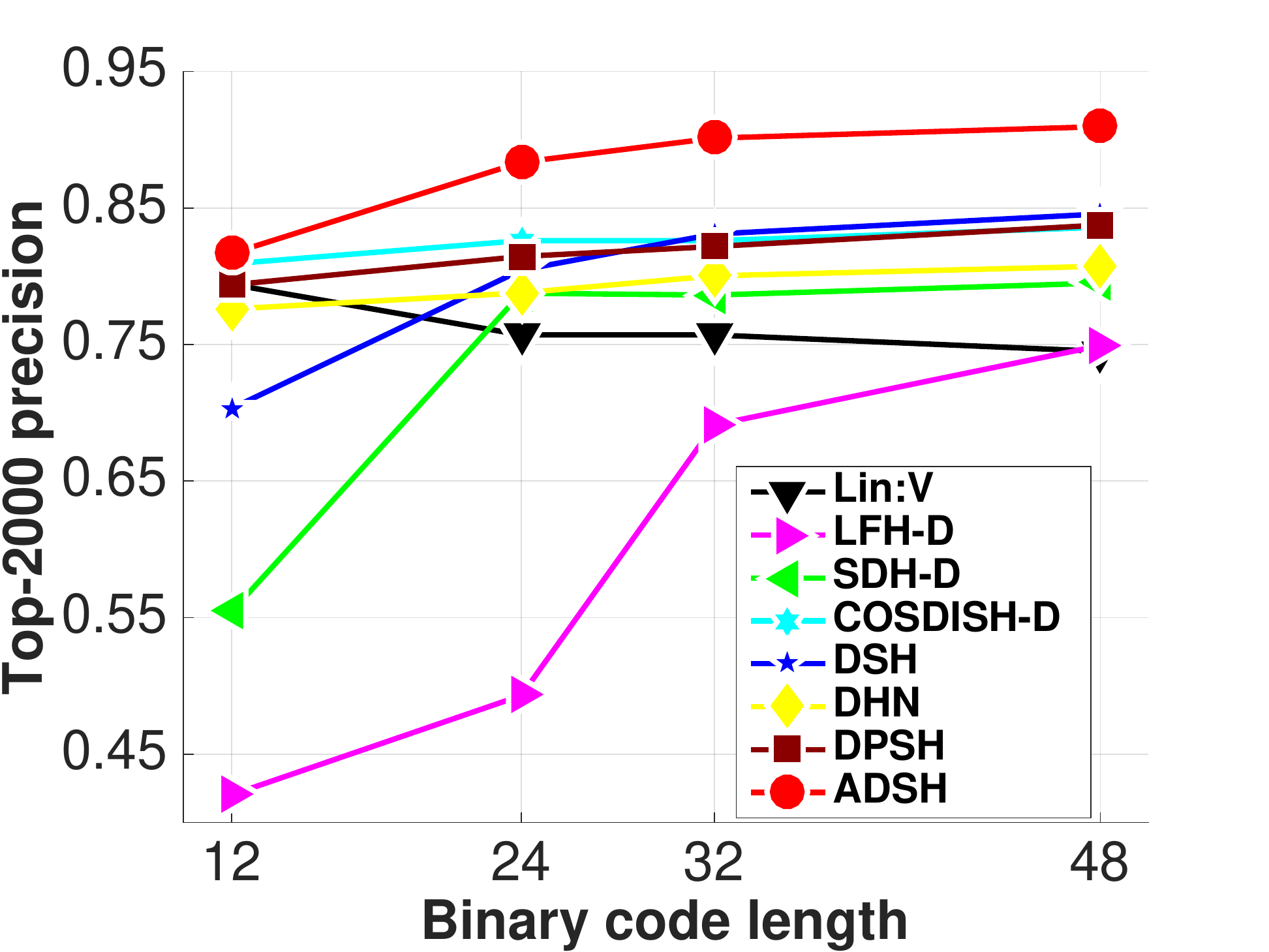}\\
    (a) CIFAR-10
\end{minipage} &
\begin{minipage}{0.49\linewidth}\centering
    \includegraphics[width=1\textwidth]{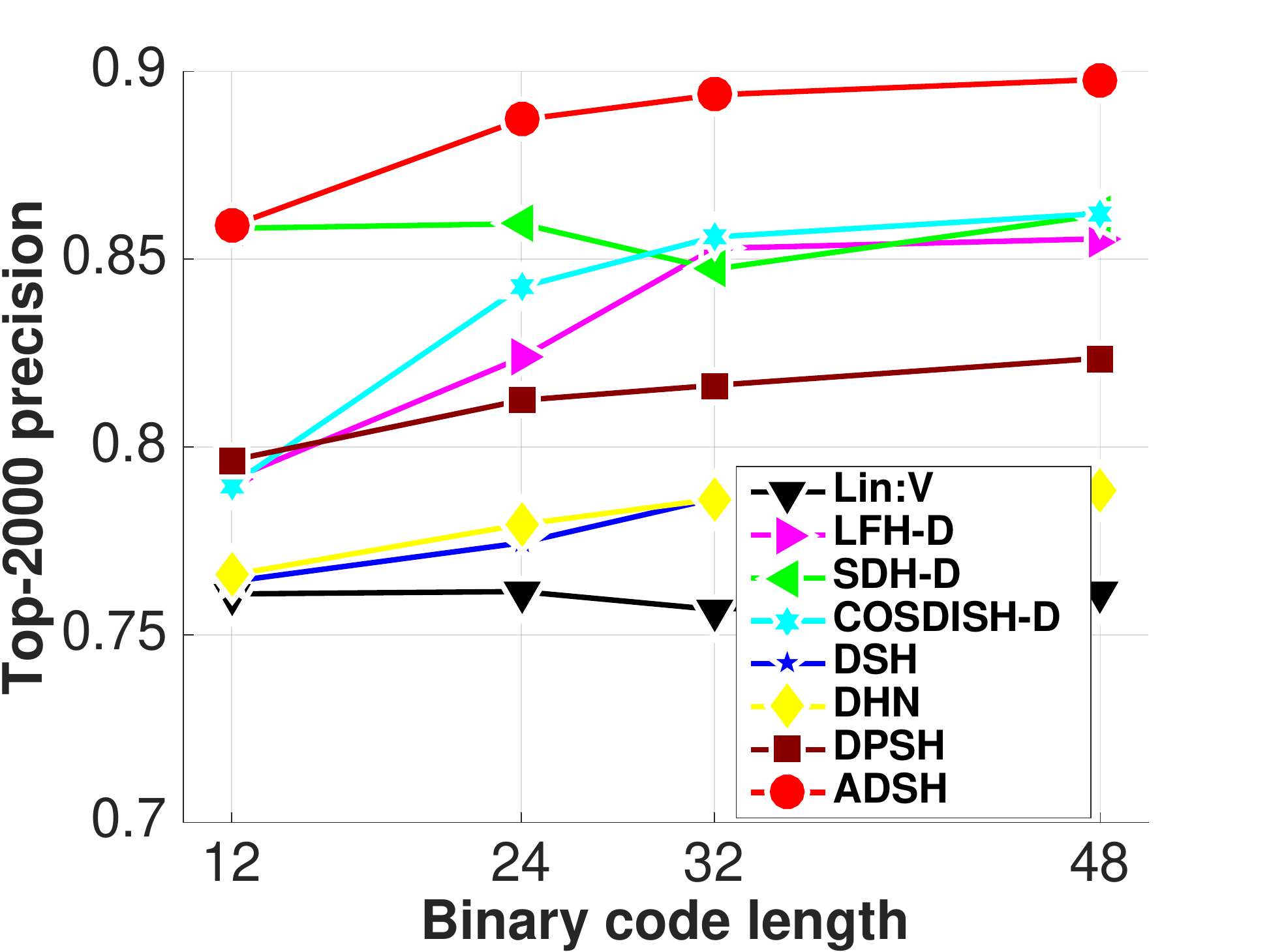}\\
    (b) NUS-WIDE
\end{minipage}
\end{tabular}
  \caption{Top-2000 precision on two datasets.}\label{fig:top2kpre}
\endminipage\hfill
\minipage{0.49\textwidth}
\begin{tabular}{c@{}@{}c@{}@{}c@{}@{}c}
\begin{minipage}{0.49\linewidth}\centering
    \includegraphics[width=1\textwidth]{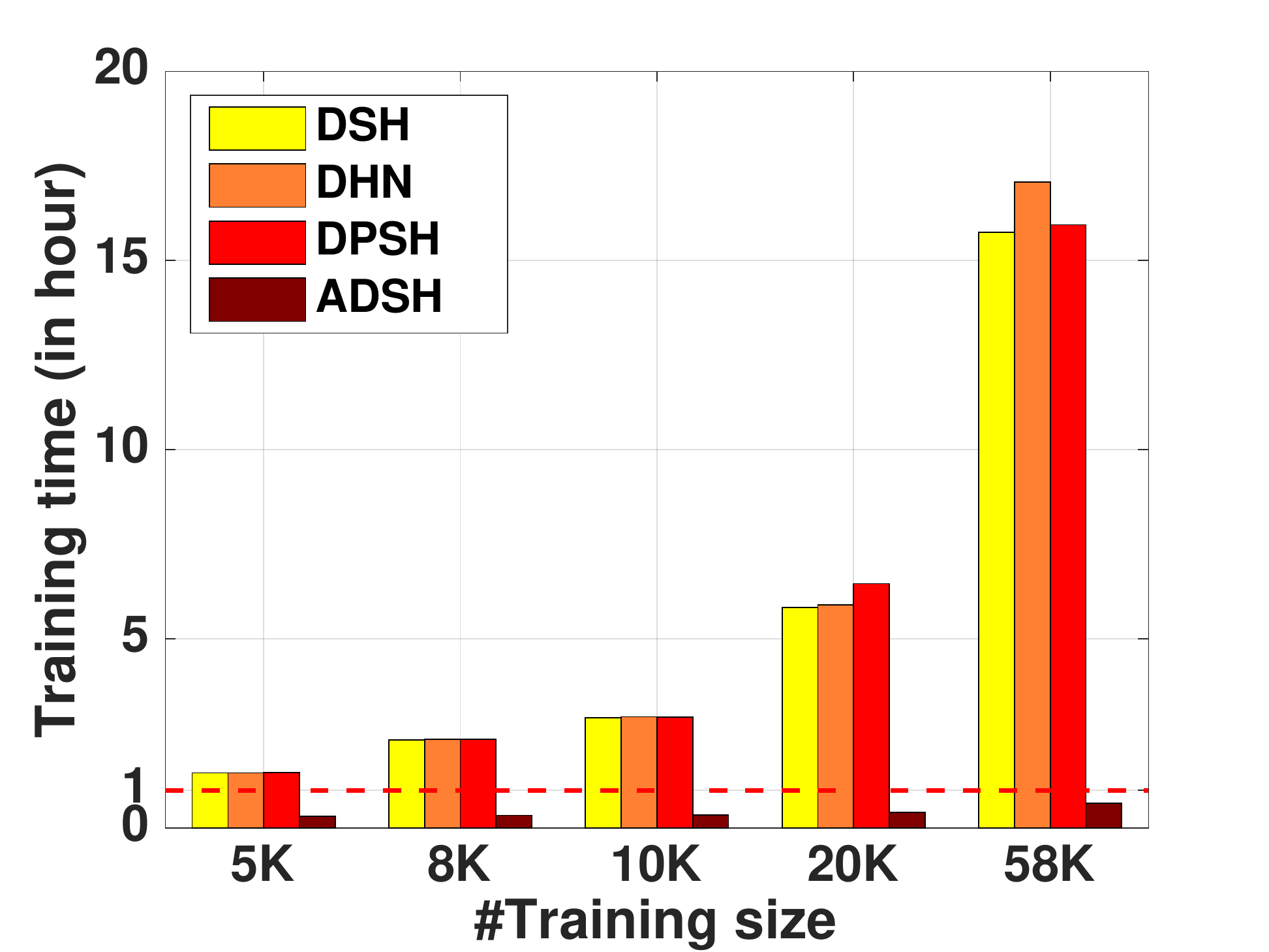}\\
    (a) CIFAR-10
\end{minipage} &
\begin{minipage}{0.49\linewidth}\centering
    \includegraphics[width=1\textwidth]{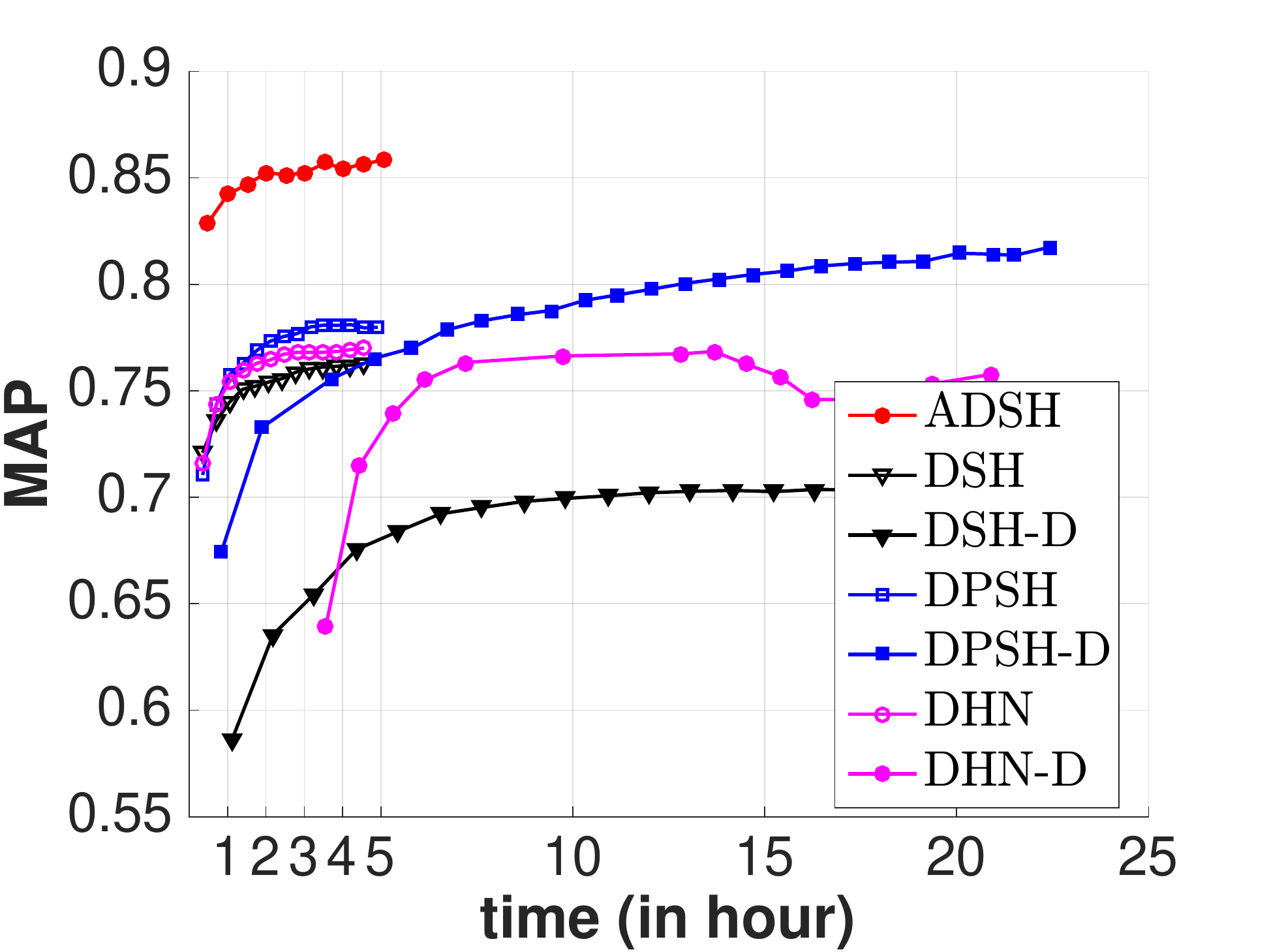}\\
    (b) NUS-WIDE
\end{minipage}
\end{tabular}
  \caption{Training time on two datasets.}\label{fig:trainingtime}
\endminipage\hfill
\end{figure}

We compare the training time of ADSH to that of other deep supervised hashing baselines on CIFAR-10 by changing the number of training points. The results are shown in Figure~\ref{fig:trainingtime} (a). We can find that ADSH is much faster than other deep hashing methods in all cases. We can also find that as the number of training points increases, the computation cost for traditional deep hashing baselines increases dramatically. On the contrary, the computation cost for ADSH increases slowly as the size of training set increases. For example, we can find that ADSH with 58,000 training points is still much faster than all the other deep baselines with 5,000 training points. Hence, we can find that ADSH can achieve higher accuracy with much faster training speed.

Furthermore, we compare ADSH to deep hashing baselines by adopting the whole database as the training set on NUS-WIDE with 12 bits. The results are shown in Figure~\ref{fig:trainingtime}~(b). Here, DSH, DHN and DPSH denote the deep hashing baselines with 10,500 sampled points for training. DSH-D, DHN-D and DPSH-D denote the counterparts of the corresponding deep hashing baselines which adopt the whole database for training. We can find that to achieve similar accuracy~(MAP), DSH, DHN and DPSH need much less time than their counterparts with whole database for training. Moreover, if the whole database is used for training, it need more than 10 hours for most baselines to converge for the case of 12 bits. For longer code with more bits, the time cost will be even higher. Hence, we have to sample a subset for training. From Figure~\ref{fig:trainingtime}~(b), we can also find that to achieve similar accuracy, our ADSH is much faster than all the baselines, either with sampled training points or with the whole database. In addition, ADSH can achieve a higher accuracy than all baselines with much less time. 

\section{Conclusion}
\label{sec:Con}

In this paper, we propose a novel deep supervised hashing method, called ADSH, for large-scale nearest neighbor search. To the best of our knowledge, this is the first work to adopt an asymmetric strategy for deep supervised hashing. Experiments show that ADSH can achieve the state-of-the-art performance in real applications.



{\small
\bibliography{reference}
\bibliographystyle{abbrv}
}
\appendix
\section{Precision-Recall Curve}
Typically, there are two retrieval protocols for hashing-base search~\cite{DBLP:conf/nips/LiuMKC14}. One is called Hamming ranking, and the other is called hash lookup. The MAP metric and top-k precision measure the accuracy of Hamming ranking. For the hash lookup protocol, precision-recall curve is usually adopted for measuring the accuracy.

\begin{figure}[!htb]
\centering
\small
\begin{tabular}{c@{}@{}c@{}@{}c@{}@{}c}
\begin{minipage}{0.24\linewidth}\centering
    \includegraphics[width=1\textwidth]{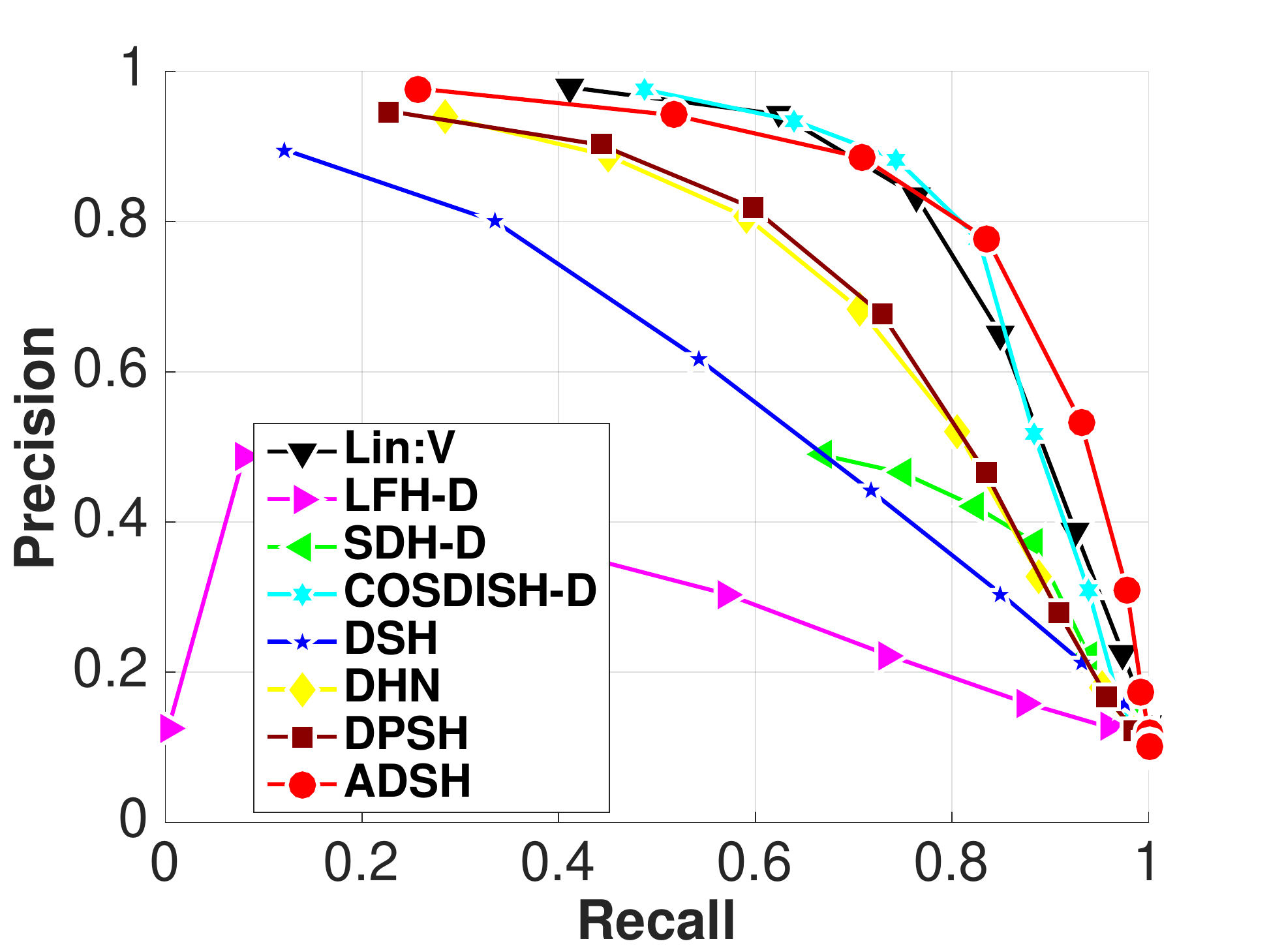}\\
    (a) 12 bits @CIFAR-10
\end{minipage} &
\begin{minipage}{0.24\linewidth}\centering
    \includegraphics[width=1\textwidth]{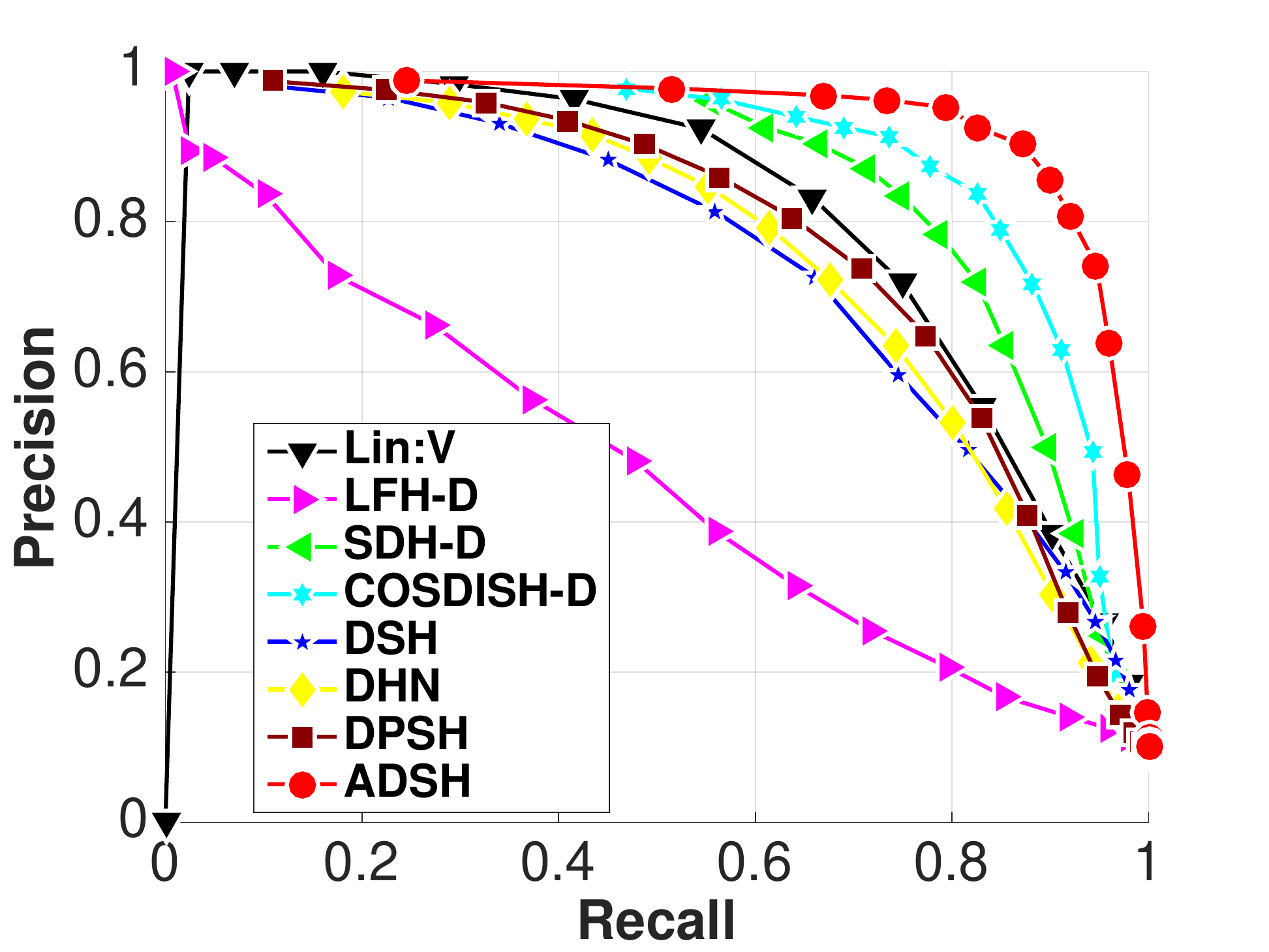}\\
    (b) 24 bits @CIFAR-10
\end{minipage} &
\begin{minipage}{0.24\linewidth}\centering
    \includegraphics[width=1\textwidth]{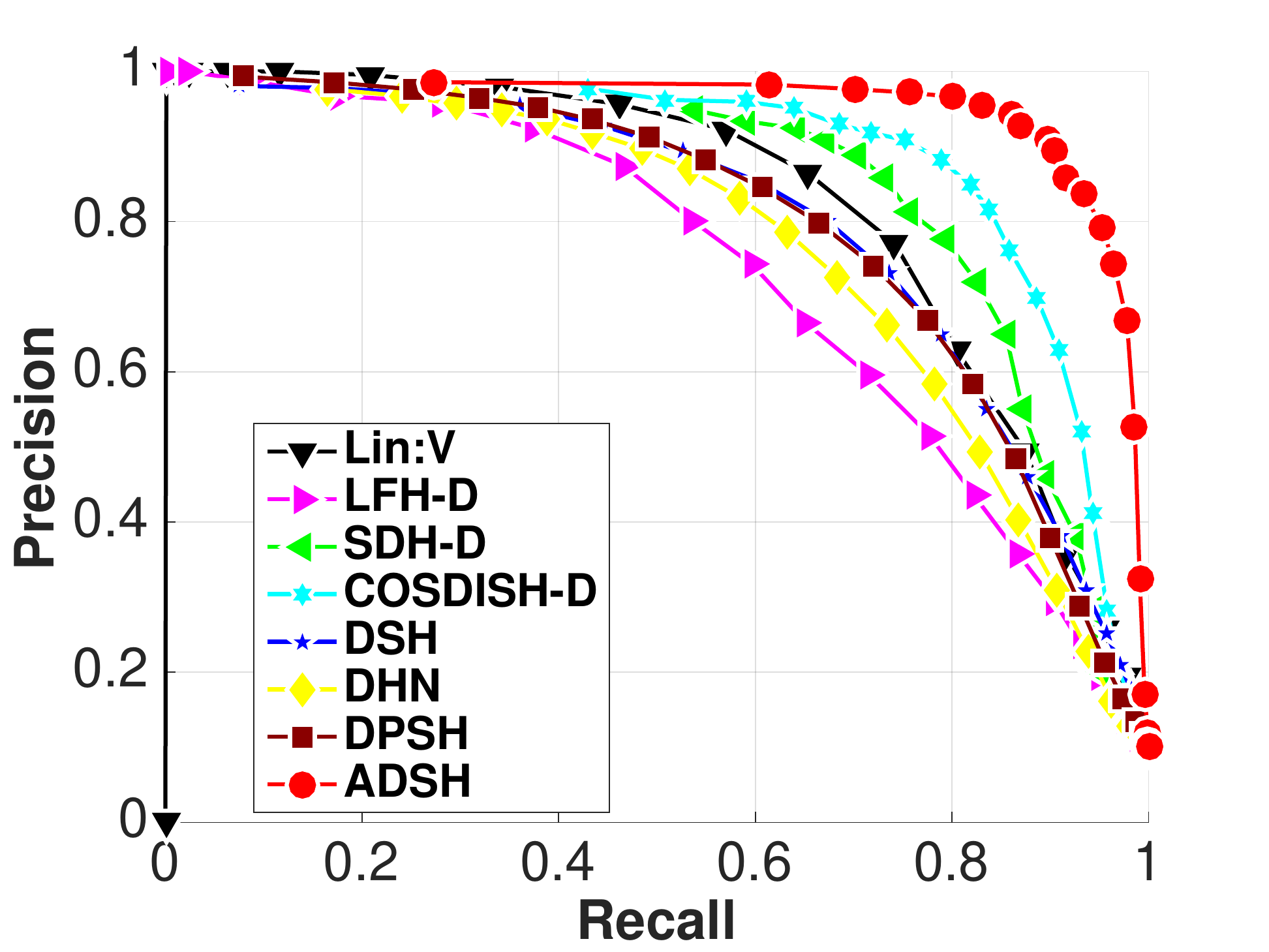}\\
    (c) 32 bits @CIFAR-10
\end{minipage} &
\begin{minipage}{0.24\linewidth}\centering
    \includegraphics[width=1\textwidth]{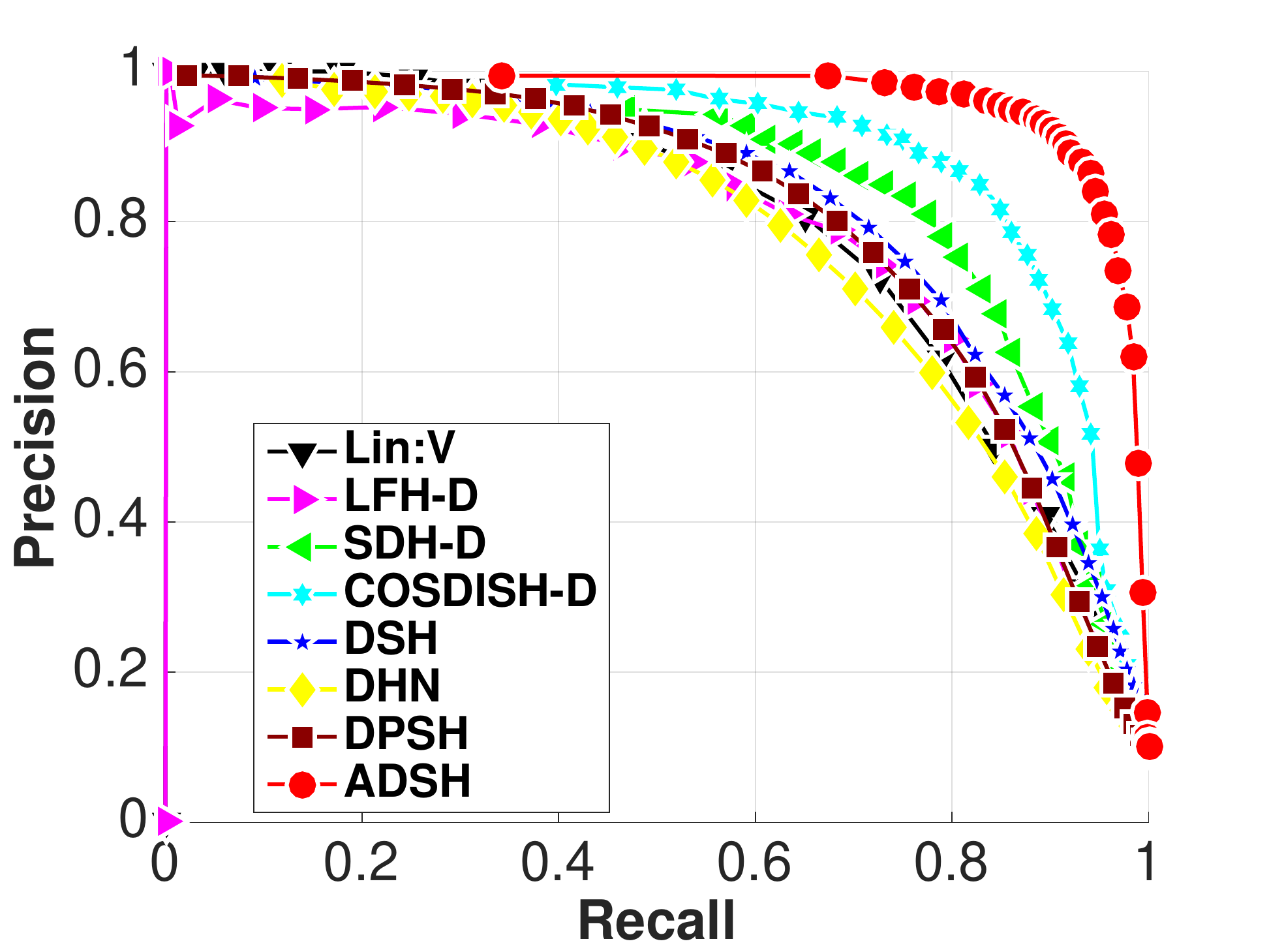}\\
    (d) 48 bits @CIFAR-10
\end{minipage} \\
\begin{minipage}{0.24\linewidth}\centering
    \includegraphics[width=1\textwidth]{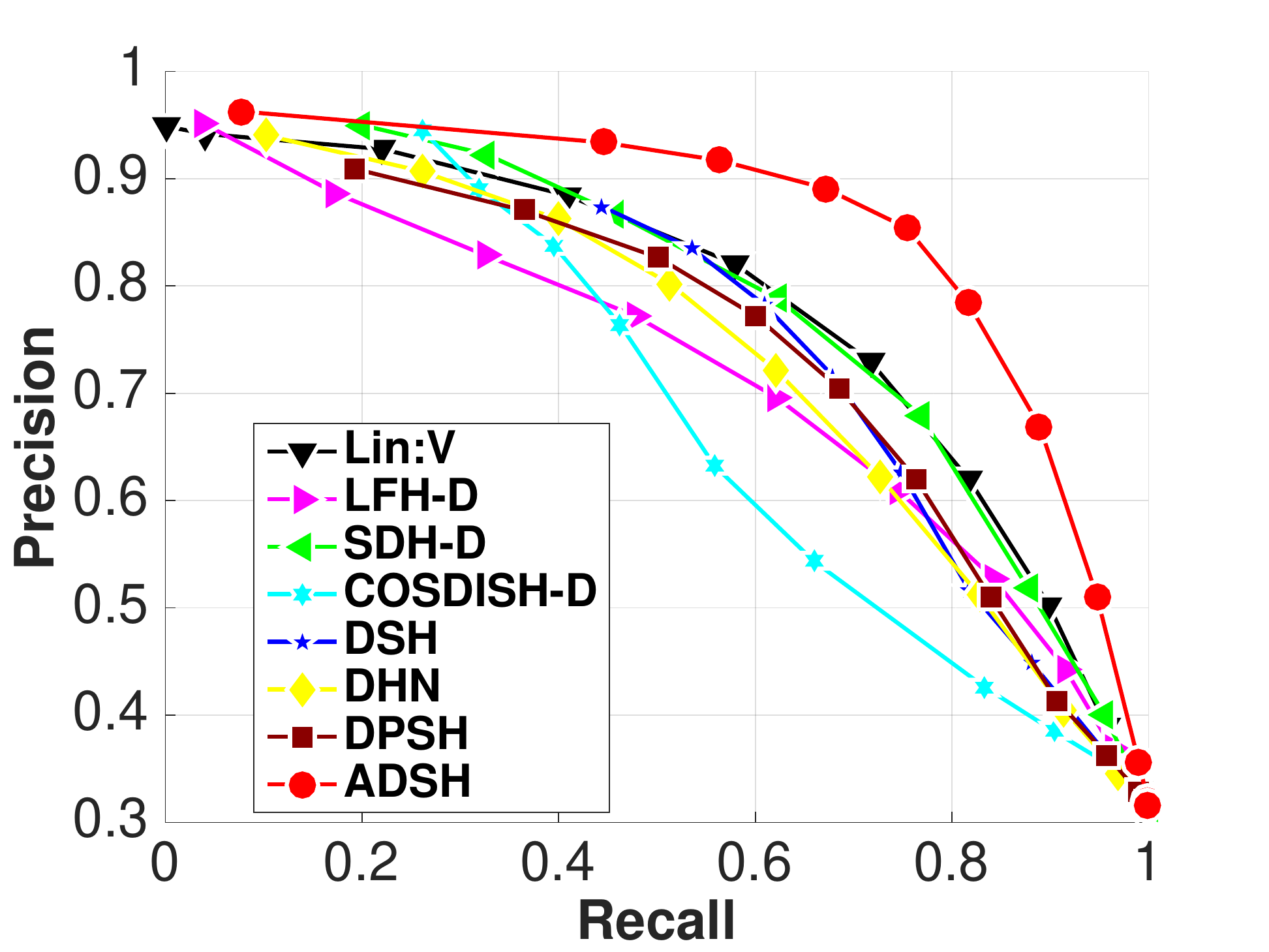}\\
    (e) 12 bits @NUS-WIDE
\end{minipage} &
\begin{minipage}{0.24\linewidth}\centering
    \includegraphics[width=1\textwidth]{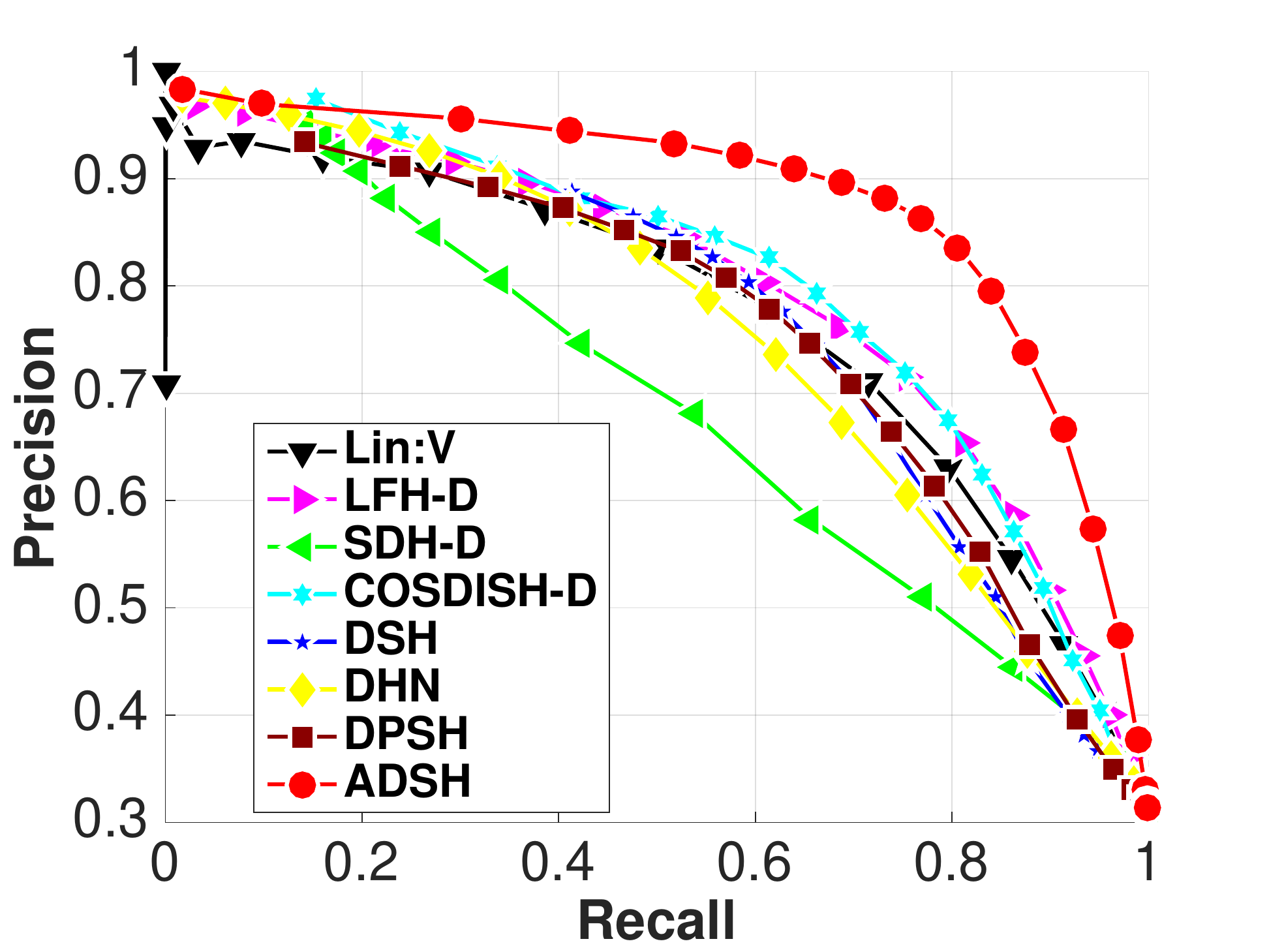}\\
    (f) 24 bits @NUS-WIDE
\end{minipage} &
\begin{minipage}{0.24\linewidth}\centering
    \includegraphics[width=1\textwidth]{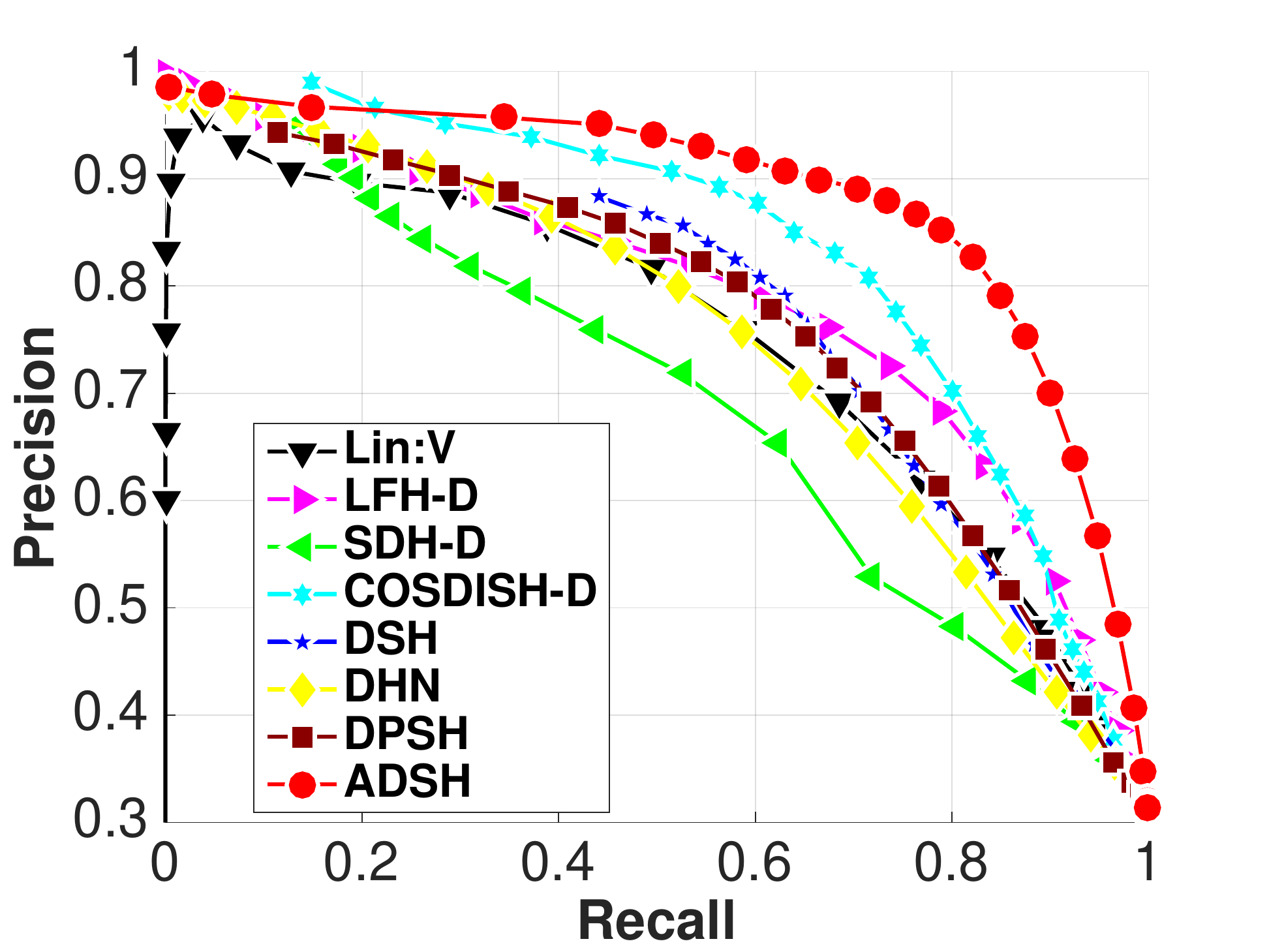}\\
    (g) 32 bits @NUS-WIDE
\end{minipage} &
\begin{minipage}{0.24\linewidth}\centering
    \includegraphics[width=1\textwidth]{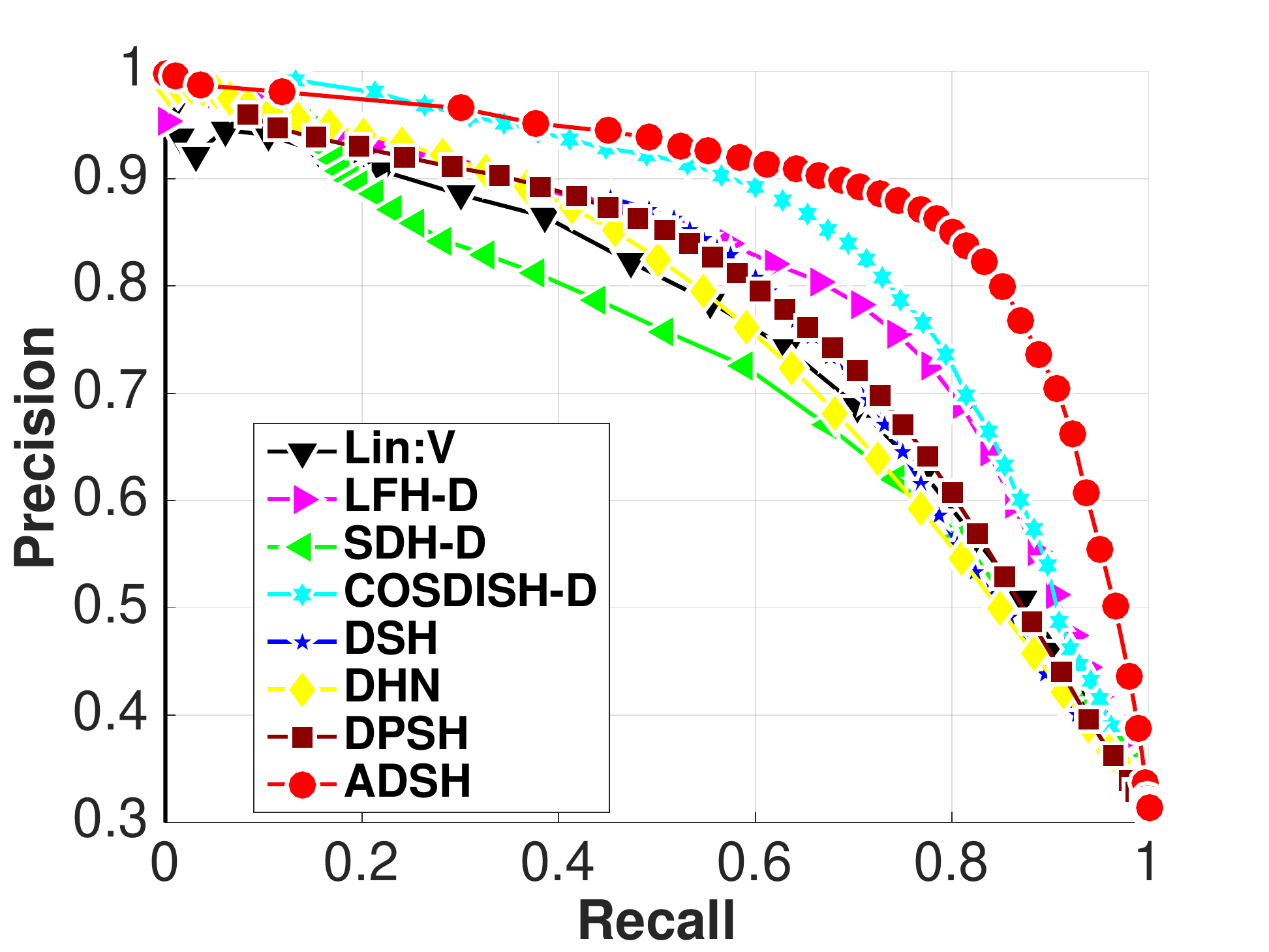}\\
    (h) 48 bits @NUS-WIDE
\end{minipage}
\end{tabular}
\caption{Precision recall curve on two dataset.}
\label{fig:prNUS-WIDE}
\end{figure}

We report precision-recall curve on two datasets in Figure~\ref{fig:prNUS-WIDE}. Each marked point in the curve corresponds to a Hamming radius to the binary hash code of the query. We can find that in most cases ADSH can achieve the best performance on both datasets.

\section{The Effectiveness of CNN Network Structure}
Since we change the network structure of DHN and DSH for fair comparison, we compare the retrieval accuracy on CIFAR-10 dataset with different network structures. The results are reported in Table~\ref{table:mapNetWork}.

We can see that the methods with CNN-F can achieve higher accuracy than the methods with the original network structures. For fair comparison, we utilize CNN-F as the network structure for all deep hashing methods.
\section{Sensitity to the Hyper-parameters}
\begin{figure*}[!htb]
\centering
\begin{tabular}{c@{}@{}c@{}@{}c@{}@{}c}
\begin{minipage}{0.4\linewidth}\centering
    \includegraphics[width=1\textwidth]{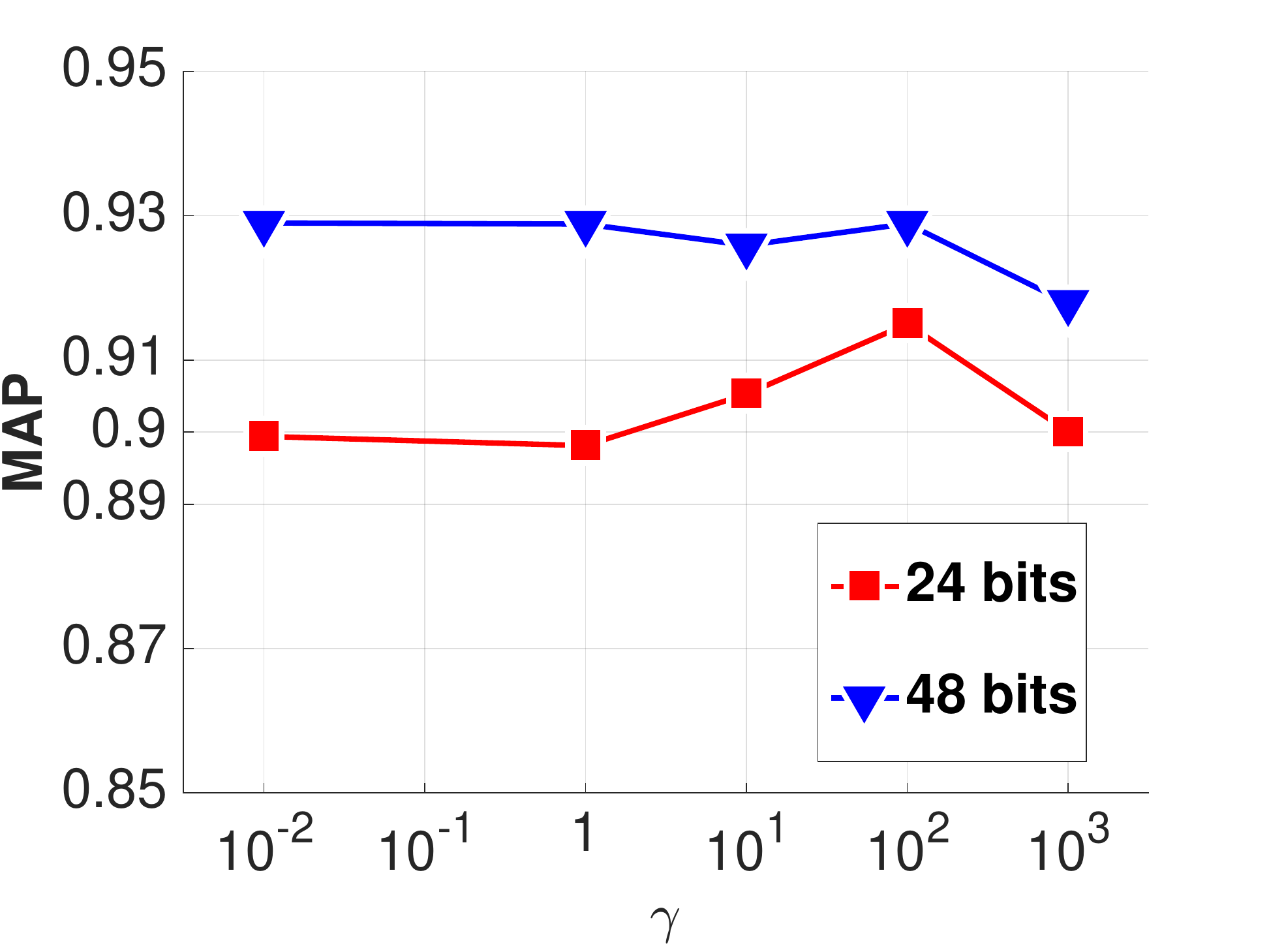}\\
    (a) $\gamma$
\end{minipage} &
\begin{minipage}{0.4\linewidth}\centering
    \includegraphics[width=1\textwidth]{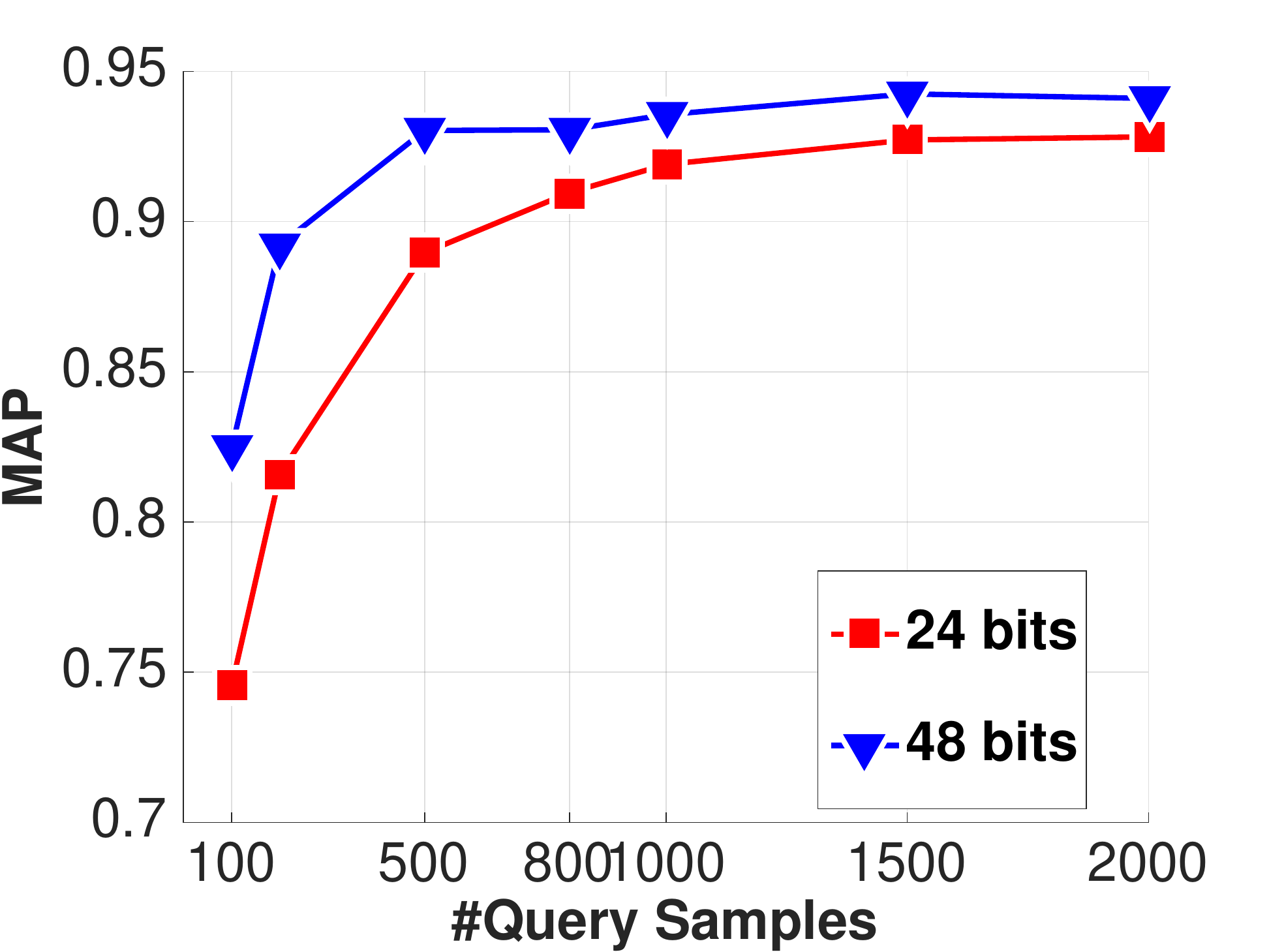}\\
    (b) \#query points
\end{minipage}
\end{tabular}\vspace{5pt}
\caption{Hyper-parameters on CIFAR-10 datasets.}
\label{fig:hyper-param}
\end{figure*}

Figure~\ref{fig:hyper-param} presents the effect of the hyper-parameter $\gamma$ and the number of query points~($m$ or $\vert\Omega\vert$) for ADSH on CIFAR-10, with binary code length being 24 bits and 48 bits. From Figure~\ref{fig:hyper-param}~(a), we can see that ADSH is not sensitive to $\gamma$ in a large range with $10^{-2}<\gamma<10^3$.

Figure~\ref{fig:hyper-param}~(b) presents the MAP results for different number of sampled query points~($m$) for ADSH on CIFAR-10. We can find that better retrieval accuracy can be achieved with larger number of sampled query points. Because larger number of sampled query points will result in higher computation cost, in our experiments we select a suitable number to get a tradeoff between retrieval accuracy and computation cost. By choosing $m=1000$, our ADSH can significantly outperform all other deep supervised hashing baselines in terms of both accuracy and efficiency.

\begin{table}[t]
\centering
\small
\caption{MAP on CIFAR-10 dataset.}
\label{table:mapNetWork}
\begin{tabular}{|c|c|c|c|c|c|c|c|}
 \hline
 Method & Network & 12 bits & 24 bits & 32 bits & 48 bits\\
 \hline
 \hline
 DHN & AlexNet & {0.5512} & {0.6091} & {0.6114} & {0.6433}\\
 \hline
 DHN & CNN-F & {0.5978} & {0.6399} & {0.6471} & {0.6870} \\
 \hline
 DSH & Proposed & {0.6157} & {0.6512} & {0.6607} & {0.6755}\\
 \hline
 DSH & CNN-F & {0.6036} & {0.6829} & {0.7245} & {0.7657}\\
 \hline
 \end{tabular}
\end{table}
\section{Case Study}
We randomly sample some test data points and show their top-10 returned results from retrieval set for each test point as a case study on CIFAR-10. For ADSH and the baselines, we use the same test points to show the results.

\begin{figure*}[!htb]
\centering
\begin{tabular}{c@{}@{}c@{}@{}c@{}@{}c}
\begin{minipage}{0.48\linewidth}\centering
    \includegraphics[width=1\textwidth]{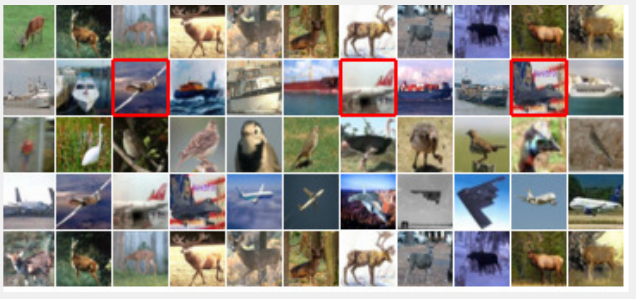}\\
    (a) ADSH\vspace{8pt}
\end{minipage} &
\begin{minipage}{0.48\linewidth}\centering
    \includegraphics[width=1\textwidth]{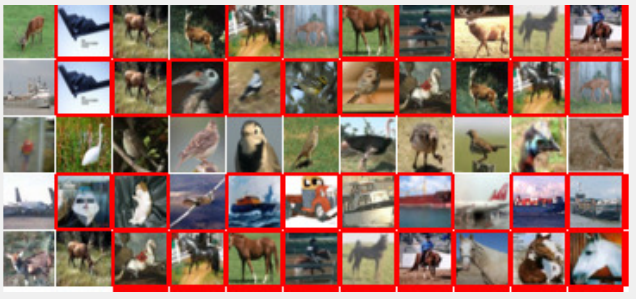}\\
    (b) LFH-D\vspace{8pt}
\end{minipage} \\
\begin{minipage}{0.48\linewidth}\centering
    \includegraphics[width=1\textwidth]{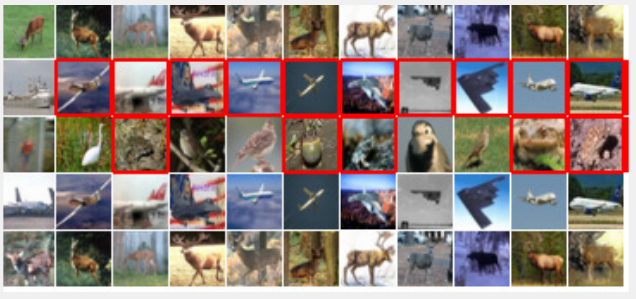}\\
    (c) SDH-D\vspace{8pt}
\end{minipage} &
\begin{minipage}{0.48\linewidth}\centering
    \includegraphics[width=1\textwidth]{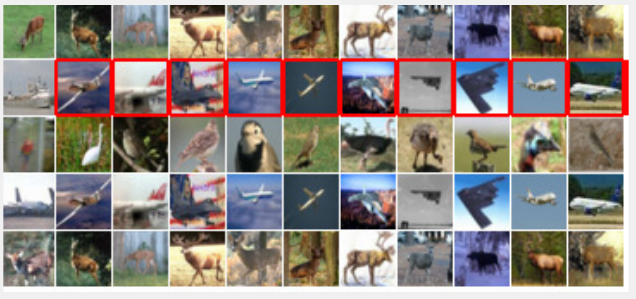}\\
    (d) COSDISH-D\vspace{8pt}
\end{minipage}
\end{tabular}\vspace{5pt}
\caption{Case study with 12 bits on CIFAR-10 dataset. For each sub-figure, the first column denotes the test images. The following 10 columns denote the top-10 returned samples. A red rectangle is used to indicate that the returned image is not the ground-truth neighbor of the corresponding test image.}
\label{fig:caseStudy}
\end{figure*}

Figure~\ref{fig:caseStudy} shows the returned results with 12 bits. We use a red rectangle to indicate that the returned image is not the ground-truth neighbor of the corresponding test image. By comparing ADSH to three best baselines, we can see that ADSH can significantly outperform other baselines.

\end{document}